%% file: main.tex
\documentclass[letterpaper, 10 pt, conference]{ieeeconf}  

\IEEEoverridecommandlockouts                              

\usepackage{graphics} 
\usepackage{epsfig} 
\usepackage{times} 
\usepackage{amsmath} 
\usepackage{amssymb}  
\usepackage{xcolor}
\usepackage{multirow}
\usepackage{cite}
\usepackage{hyperref} 
\usepackage{amsfonts}
\usepackage{algorithmic}
\usepackage{graphicx}
\usepackage{textcomp}
\usepackage{caption}
\usepackage{array}  
\let\labelindent\relax
\usepackage{enumitem}
\usepackage{balance}      
\usepackage{placeins}     
\usepackage[table]{xcolor}
\setlist[itemize]{noitemsep, nolistsep,leftmargin=*}

\newcommand{\ourmethod}{\texttt{ResMimic} }

\title{
\textbf{ResMimic}: From General Motion Tracking to Humanoid Whole-body Loco-Manipulation via Residual Learning
}

\author{ 
 Siheng Zhao$^{12\text{\textsection}}$\quad Yanjie Ze$^{13\text{\textsection}}$ \quad Yue Wang$^{2}$ \quad C. Karen Liu$^{13\dag{}}$ \\ Pieter Abbeel$^{14\dag{}}$\quad Guanya Shi$^{15\dag{}}$\quad Rocky Duan$^{1\dag{}}$\\
$^1$Amazon FAR (Frontier AI \& Robotics)\quad $^2$USC\quad  $^3$Stanford University\quad  $^4$UC Berkeley \quad  $^5$CMU\quad 
\\
\\
\href{https://resmimic.github.io/}{https://resmimic.github.io/}
}

\begin{document}
\twocolumn[{%
\renewcommand\twocolumn[1][]{#1}%
\maketitle
\vspace{-0.2in}
\begin{center}
    \centering
    \captionsetup{type=figure}
   \includegraphics[width=1.0\textwidth]{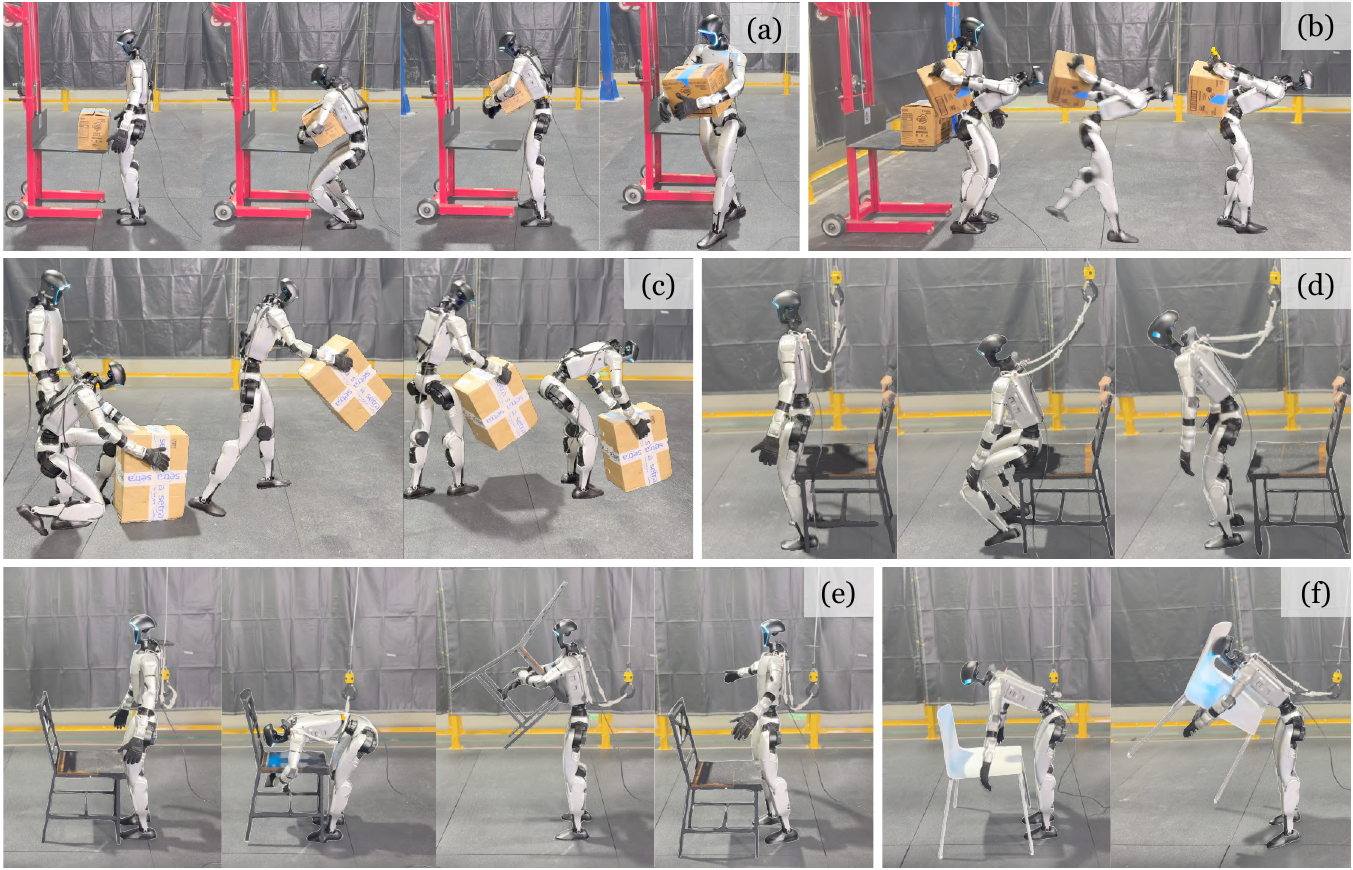}
     \vspace{-0.2in}
    \caption{We deploy \ourmethod on a Unitree G1 humanoid to demonstrate diverse whole-body loco-manipulation capabilities. (\textbf{a}) Carrying 4.5kg heavy payloads via whole-body contact; (\textbf{b}, \textbf{c}) expressive motion while carrying box; (\textbf{d}) general humanoid object interaction beyond manipulation; (\textbf{e}, \textbf{f}) carrying irregularly shaped heavy objects with instance-level generalization. 
}
    \label{fig/teaser}
\end{center}
\vspace{-0.05in}
}]

\thispagestyle{empty}
\pagestyle{empty}

\renewcommand{\thefootnote}{\fnsymbol{footnote}}
\begin{abstract}
\footnotetext[4]{Work done during the internship of Siheng Zhao and Yanjie Ze at Amazon FAR (Frontier AI \& Robotics)~~~$^{\dag{}}$FAR Co-Lead}
Humanoid whole-body loco-manipulation promises transformative capabilities for daily service and warehouse tasks. While recent advances in general motion tracking (GMT) have enabled humanoids to reproduce diverse human motions, these policies lack the precision and object awareness required for loco-manipulation.
To this end, we introduce \texttt{ResMimic}, a two-stage residual learning framework for precise and expressive humanoid control from human motion data. First, a GMT policy, trained on large-scale human-only motion, serves as a task-agnostic base for generating human-like whole-body movements. An efficient but precise residual policy is then learned to refine the GMT outputs to improve locomotion and incorporate object interaction. To further facilitate efficient training, we design (i) a point-cloud–based object tracking reward for smoother optimization, (ii) a contact reward that encourages accurate humanoid body–object interactions, and (iii) a curriculum-based virtual object controller to stabilize early training.
We evaluate \texttt{ResMimic} in both simulation and on a real Unitree G1 humanoid. Results show substantial gains in task success, training efficiency, and robustness over strong baselines. 
Videos are available at \href{resmimic.github.io}{resmimic.github.io}.
\end{abstract}
\renewcommand{\thefootnote}{\arabic{footnote}}

\input{sections/intro}
\input{sections/related}
\input{sections/method}

\input{sections/exp}

\input{sections/conclusion}






\FloatBarrier

\balance
\bibliographystyle{IEEEtran}
\bibliography{main}

\end{document}

%% file: sections/intro.tex
\section{INTRODUCTION}

Humanoid robots have been showing  promising applications thanks to their flexibility. Unlike quadrupeds or wheeled manipulators, they can perform expressive whole-body loco-manipulation, combining locomotion and manipulation in a coordinated, human-like manner. This unique capability opens the door to transformative applications in daily services and industrial operations, while leveraging existing human-centric infrastructure.

However, realizing precise and expressive humanoid loco-manipulation remains a fundamental challenge to humanoids. Compared to locomotion~\cite{TokenHumanoid2024, HumanoidTransformer2023} or tabletop manipulation~\cite{okami2024,ze2024humanoid_manipulation}, loco-manipulation demands higher precision, rich whole-body contacts, and data that is not easily accessible at scale. While direct imitation of human motions~\cite{2018-TOG-deepMimic} is attractive—since humans naturally perform coordinated full-body control—retargeting human demonstrations to humanoid robots introduces significant embodiment gap: contact locations and relative object pose in human demonstrations often fail to translate, leading to floating contacts or penetrations, as shown in Figure~\ref{fig/imperfet_data}.

Recent progress on general motion tracking (GMT) policies~\cite{ze2025twist,chen2025gmt}, trained on large-scale human-only datasets~\cite{AMASS:ICCV:2019, li2023objectmotionguidedhuman}, shows that humanoids can reproduce diverse human motions with high fidelity. However, these policies are unaware of manipulated objects. On the other hand, existing humanoid loco-manipulation approaches rely on highly task-specific designs, such as stage-wise controllers~\cite{dao2023simtoreallearninghumanoidbox} or handcrafted data pipelines~\cite{liu2024opt2skill}, which limit scalability and generality. As a result, there is still no unified, efficient, and precise framework for humanoid loco-manipulation yet.

In parallel, breakthroughs in foundation models~\cite{openai2024gpt4technicalreport} have demonstrated the power of pre-training on large-scale data followed by post-training. In robotics manipulation, models like $\pi_0$~\cite{black2024pi0visionlanguageactionflowmodel} and LBM~\cite{trilbmteam2025carefulexaminationlargebehavior} highlight that this paradigm resolves data scarcity while improving robustness and generalization. In graphics, pretrained latent spaces are similarly leveraged for downstream generation and editing~\cite{roich2021pivotaltuninglatentbasedediting, gal2021stylegannadaclipguideddomainadaptation}. However, in humanoid whole-body control, such a powerful pretrain–finetune paradigm remains largely unexplored.

In this paper, our key insight is that while diverse human motions can be effectively mimiced by a pre-trained GMT policy, object-centric loco-manipulation requires task-specific corrections. Importantly, many whole-body motions—such as balance, stepping, or reaching—are shared across tasks, while only the fine-grained object interaction requires adaptation. This motivates a residual learning paradigm, where a stable motion prior is augmented with lightweight task-specific adjustments. 

To this end, we propose \ourmethod, a two-stage residual learning framework for humanoid loco-manipulation. First, a GMT policy is trained on large-scale motion capture data to reproduce diverse human motions, serving as a robust prior for human-like whole-body behavior. Second, a task-specific residual policy is trained efficiently to condition on object reference trajectory, outputting corrective actions that refine the GMT policy and enable precise object manipulation. This decoupling alleviates the need for per-task reward engineering, improves data efficiency, and yields a general framework applicable not only to loco-manipulation but also to locomotion enhancement.

Our contributions are four-fold:
\begin{itemize}
    \item We propose a two-stage residual learning framework that combines pre-trained GMT with task-specific corrections, enabling efficient, precise humanoid loco-manipulation.

\item To improve training efficiency and sim-to-real transfer, we propose (i) a point-cloud–based object tracking reward for smoother optimization, (ii) a contact reward that explicitly guides humanoid–object contacts, and (iii) a virtual object controller that provides a curriculum-based warm start.

\item We conduct extensive evaluations in simulation and real world. Results demonstrate significant improvements in human motion tracking, object motion tracking, task success rate, training efficiency, robustness, and generalization on challenging loco-manipulation tasks.

\item  To accelerate research on humanoid loco-manipulation, we will release our GPU-accelerated simulation infrastructure, sim-to-sim evaluation prototype, and motion data.
\end{itemize}

%% file: sections/related.tex
\section{RELATED WORK}
\begin{figure}[t]
\centering
\includegraphics[width=0.99\linewidth]{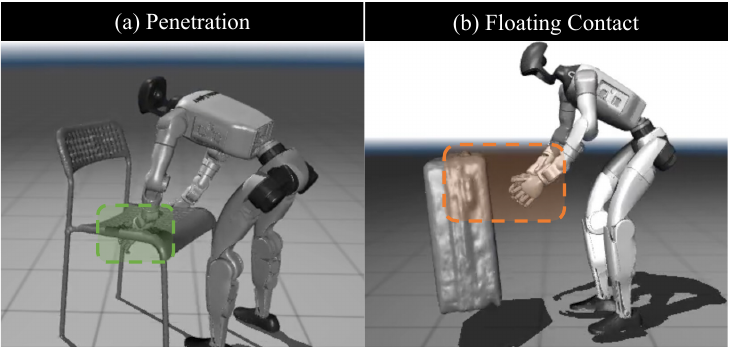}
\caption{Visualization of imperfect humanoid–object interaction data caused by the embodiment gap during retargeting: {\textbf{(a)}} hand–chair penetration; {\textbf{(b)}} hand–box floating contact.}
\label{fig/imperfet_data}
\vspace{-0.2in}
\end{figure}

\subsection{Learning-Based Humanoid Control}
Reinforcement learning (RL) enables real-time whole-body control of humanoids by training directly via interactions with the environment. 
However, such training typically suffers from low data efficiency and requires substantial effort in task-specific reward design~\cite{sferrazza2024humanoidbenchsimulatedhumanoidbenchmark}. As a result, most prior work has focused primarily on locomotion~\cite{scirobotics,zhuang2024humanoid} rather than versatile whole-body control, or is limited in specific task such as getting up~\cite{huang2025host} and keeping balance~\cite{zhang2025hublearningextremehumanoid}.

To enable more general control, learning from human motions has emerged as a promising direction~\cite{2018-TOG-deepMimic}. These methods typically rely on kinematic retargeting to map human motions to humanoids, addressing the large embodiment gap. It enables accurate tracking of individual motions~\cite{he2025asap,liao2025beyondmimicmotiontrackingversatile,mao2024learningmassivehumanvideos} or aims for versatile general motion tracking~\cite{cheng2024express,he2024omnih2o,ze2025twist,chen2025gmt}. While these methods demonstrate whole-body control ability, they remain limited in their ability to interact with physical world. VideoMimic~\cite{videomimic} makes notable progress by not only tracking human motion but also reconstructing the environment, enabling contextual motions such as sitting on a chair. However, it still interacts only with static environments and does not extend to dynamic object interactions—an essential step toward real-world utility. 
Different from these works, our work focuses on dynamic loco-manipulation.

\begin{figure*}[t]
\centering
\includegraphics[width=1\textwidth]{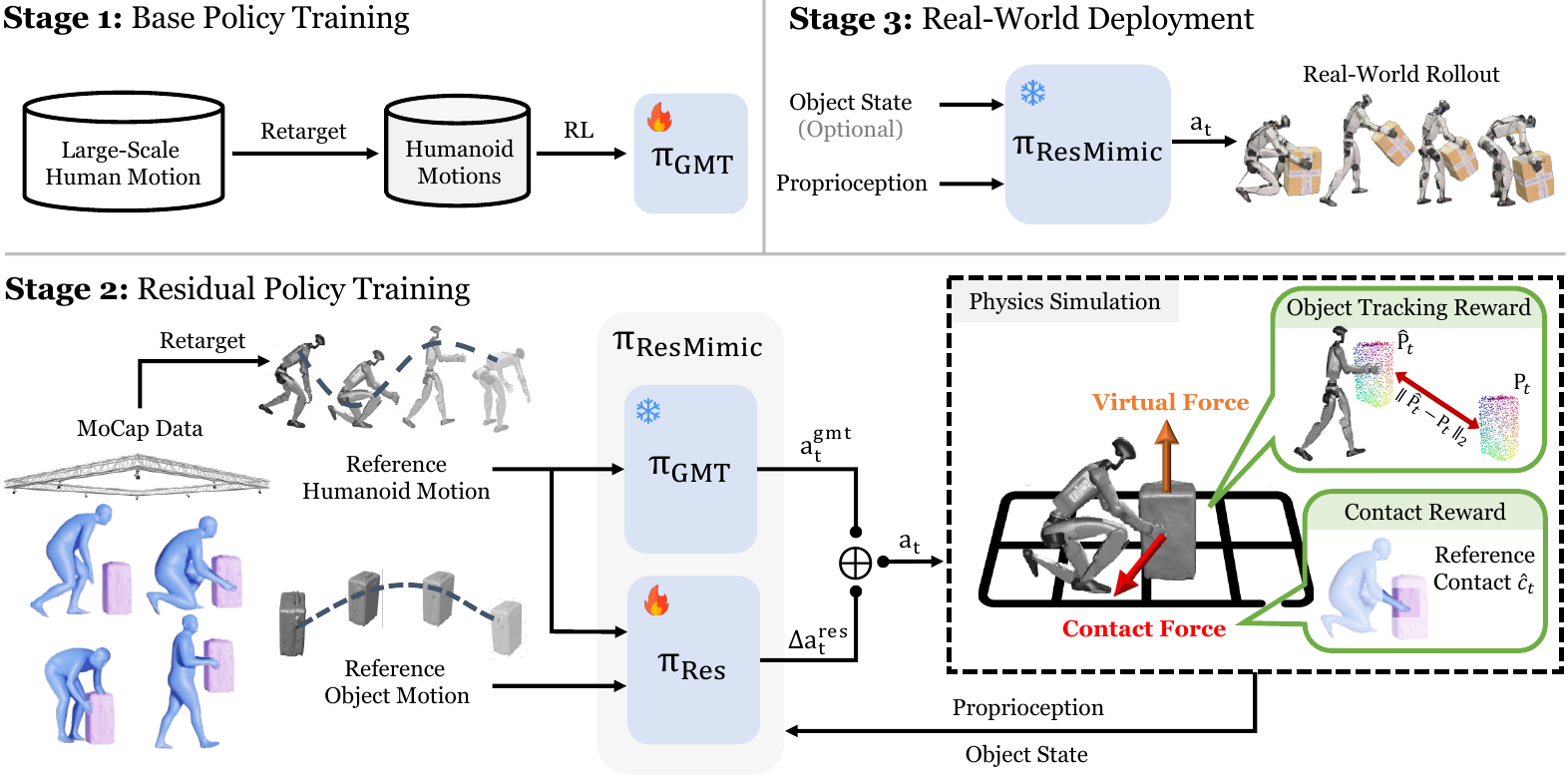}
\caption{Overview of \ourmethod: (1) A general motion tracking policy is trained on large-scale human motion data to serve as base policy. (2) A task-specific residual policy is efficiently trained with \textcolor{orange}{virtual force}, \textcolor{green}{object and contact reward}, to refine the base policy outputs. (3) During real-world deployment, the combined policy is employed for loco-manipulation control.
}
\vspace{-0.2in}
\label{fig/pipeline}
\end{figure*}

\subsection{Humanoid Loco-Manipulation} 
Humanoid loco-manipulation remains a particularly challenging learning problem. Recent work has demonstrated promising results by leveraging teleoperation~\cite{ ze2025twist,li2025amo,ben2024homie, zhang2025falcon}. However, these approaches lack explicit object awareness and require a human operator. Building on teleoperation, some works train autonomous imitation learning policies from collected data~\cite{fu2024humanplushumanoidshadowingimitation,ze2024humanoid_manipulation}. Nonetheless, these efforts remain restricted to tabletop manipulation with limited whole-body expressiveness—tasks that could often be achieved more effectively by dual-arm mobile manipulators.  
The most related works to ours are~\cite{dao2023simtoreallearninghumanoidbox,liu2024opt2skill}. Dao \textit{et al.} \cite{dao2023simtoreallearninghumanoidbox} propose a modular sim-to-real RL pipeline for box loco-manipulation, decomposing the task into distinct phases (e.g., walking, box-picking) with separate policies for each. Liu \textit{et al.} \cite{liu2024opt2skill} introduce an end-to-end learning pipeline using reference motions generated by task-specific trajectory optimization. However, all these approaches demonstrate loco-manipulation with limited whole-body contact (e.g., using only hands) and expressiveness, as well as rely on highly task-specific design. In contrast, our method leverages a GMT policy as prior, enabling more expressive whole-body loco-manipulation under a unified framework.  

\subsection{Residual Learning for Robotics}
Residual learning on top of predefined or learned base models has been widely adopted in robotics. 
Early works introduced residual policies to refine hand-designed policies or model predictive controllers, achieving more precise and contact-rich manipulation~\cite{johannink2018residualreinforcementlearningrobot, silver2019residualpolicylearning}. 
Building on this idea, later approaches extended residual learning to policies initialized from demonstrations~\cite{alakuijala2021residualreinforcementlearningdemonstrations, haldar2023teachrobotfishversatile}. In dexterous hand manipulation, residual policies have been used to adapt human hand motions for task-oriented control~\cite{li2025maniptransefficientdexterousbimanual,zhao2024dexh2rtaskorienteddexterousmanipulation}. Notably in humanoids, ASAP~\cite{he2025asap} leverages residual learning to compensate for dynamics mismatch between simulation and reality, enabling agile whole-body skills. Distinct from these directions, our method leverages a pre-trained general motion tracking (GMT) policy as a foundation and learns a residual policy to enable expressive whole-body loco-manipulation.

%% file: sections/method.tex
\section{METHOD}

We formulate our whole-body loco-manipulation task as a goal-conditioned RL problem within a Markov Decision Process (MDP) $\mathcal{M} = \langle \mathcal{S}, \mathcal{A}, \mathcal{T}, \mathcal{R}, \gamma \rangle$, where $\mathcal{S}$ is the state space, $\mathcal{A}$ the action space, $\mathcal{T}$ the transition dynamics, $\mathcal{R}$ the reward function, and $\gamma$ the discount factor.  
At time $t$, the state $s_t \in \mathcal{S}$ includes: (i) robot proprioception $s^r_t$, (ii) object state $s^o_t$, (iii) motion goal state $\hat{s}^{r}_t$, and (iv) object goal state $\hat{s}^{o}_t$. 
The action $a_t$ specifies target joint angles, which are executed on the robot through a PD controller.
The reward is defined as  $r_t = \mathcal{R}(s_t, a_t)$,
while the training objective is to maximize the expected cumulative discounted reward:  
$\mathbb{E}[\sum_{t=1}^T \gamma^{t-1} r_t]$.

\subsection{Two-Stage Residual Learning}
To transfer human–object interaction data into humanoid whole-body loco-manipulation policies in a task-agnostic manner, our objective is to avoid task-specific reward engineering, which may benefit individual tasks but limits generality. Instead, we propose a two-stage residual learning framework, as shown in Figure~\ref{fig/pipeline}.   \textbf{Stage I: General Motion Tracking (GMT).}  
     We first train a general human motion tracking policy $\pi_{\text{GMT}}$ as the backbone controller. Given robot proprioception $s^r_t$ and reference motion $\hat{s}^r_t$, the policy outputs a coarse action $a_t^{\text{gmt}} = \pi_{\text{GMT}}(s^r_t, \hat{s}^r_t),$
    optimized to maximize the motion tracking reward $\mathbb{E}\big[\sum_{t=1}^T \gamma^{t-1} r^m_t\big]$.  
\textbf{Stage II: Residual Refinement.}  
    Building on the pretrained GMT policy, we train an efficient and precise residual policy $\pi_{\text{Res}}$ per-task that refines the coarse action using both robot and object information:  
    $\pi_{\text{Res}}(s^r_t, s^o_t, \hat{s}^r_t, \hat{s}^o_t) = \Delta a_t^{\text{res}}.$  
    The final action is computed as  
    $a_t = a_t^{\text{gmt}} + \Delta a_t^{\text{res}},$  
    and the residual policy is optimized to maximize the combined motion and object rewards $\mathbb{E}\big[\sum_{t=1}^T \gamma^{t-1} (r^m_t + r^o_t)\big]$.  
Both stages are trained using PPO~\cite{schulman2017proximal}.

\subsection{General Motion Tracking Policy}
In this stage, our goal is to train a real-world deployable general motion tracking policy $\pi_{\text{GMT}}$ that takes only humanoid proprioception $s^r_t$ and human reference motion $\hat{s}^r_t$ as input, and outputs actions $a^{\text{gmt}}$ for the humanoid robot to mimic the reference motion. 
While \ourmethod is a general framework that can incorporate any GMT policy as the base policy, we present one specific implementation as example.

\subsubsection{Dataset} 
An important motivation of our two-stage training pipeline is to decouple human motion tracking from object interaction. The general motion tracking policy relies solely on human motion capture data, avoiding the need for costly and hard-to-obtain manipulation data.

We leverage several publicly available MoCap datasets, including AMASS~\cite{AMASS:ICCV:2019} and OMOMO~\cite{li2023objectmotionguidedhuman}, which together contain over 15,000 clips (approximately 42 hours). Motions that are impractical for our setting, such as stair climbing, are filtered out. 
After curating the human motion dataset, we apply kinematics-based motion retargeting (e.g. GMR~\cite{ze2025gmr}) to transfer human motions into a humanoid reference motion dataset $\{\hat{S}^r_i = \{\hat{s}^r_t\}_{t=1}^T\}_{i=1}^D$.

\subsubsection{Training Strategy}
To train a real-world deployable general motion tracking policy $\pi_{\text{GMT}}$, we adopt a single-stage RL framework in simulation without access to privileged information.  
The proprioceptive observation $s^r_t$ is defined as $[\theta_t, \omega_t, q_t, \dot{q}_t, a_t^{\text{hist}}]_{t-10:t}$, where $\theta_t$ is the root orientation, $\omega_t$ is the root angular velocity, $q_t \in \mathbb{R}^{29}$ is the joint position, $\dot{q}_t$ is the joint velocity, and $a_t^{\text{hist}}$ is the recent action history.  
The reference motion input $\hat{s}^r_t$ is defined as $[\hat{p}_t, \hat{\theta}_t, \hat{q}_t]_{t-10:t+10}$, where $\hat{p}_t$ is the reference root translation, $\hat{\theta}_t$ is the reference root orientation, and $\hat{q}_t$ is the reference joint position. 
To improve tracking quality, we also incorporate future reference motion into the input, enabling the policy to anticipate and plan for upcoming targets, which yields smoother tracking.

\subsubsection{Reward and Domain Randomization}
Following TWIST~\cite{ze2025twist}, the motion tracking reward $r^m_t$ is formulated as the sum of three components: (i) task rewards, (ii) penalty terms, and (iii) regularization terms. To promote robust and generalizable sim-to-real transfer, we further apply domain randomization during training. See \cite{ze2025twist} for more details. 

\subsection{Residual Refinement Policy}
Building on the pre-trained $\pi_{\text{GMT}}$, we introduce a residual policy $\pi_{\text{Res}}$ to refine the coarse actions predicted by the base policy and thereby complete the desired task.  

\subsubsection{Reference Motions}
We obtain reference trajectories using a MoCap system that simultaneously records human motion $\{\hat{h}_t\}_{t=1}^T$ and object motion $\{\hat{o}_t\}_{t=1}^T$. The human motion is retargeted to the humanoid robot using GMR~\cite{ze2025gmr}, yielding humanoid reference trajectories $\{\hat{s}^r_t = \text{GMR}(\hat{h}_t)\}_{t=1}^T$. The object motion is directly used as the reference $\hat{s}^o_t$. Together, these form the complete reference trajectory $\{(\hat{s}^r_t, \hat{s}^o_t)\}_{t=1}^T$ for training the residual policy.

\subsubsection{Training Strategy}
We adopt single-stage RL with PPO for residual learning. The residual policy $\pi_{\text{Res}}$ takes $\langle s^r_t, s^o_t, \hat{s}^r_t, \hat{s}^o_t \rangle$ as input and outputs a residual action $\Delta a_t^{\text{res}} \in \mathbb{R}^{29}$. The object state is represented as $s^o_t = [p^o_t, \theta^o_t, v^o_t, \omega^o_t]$, and the reference object trajectory as $\hat{s}^o_t = [\hat{p}^o_t, \hat{\theta}^o_t, \hat{v}^o_t, \hat{\omega}^o_t]_{t-10:t+10}$, where $p^o_t$ denotes the object root translation, $\theta^o_t$ the root orientation, $v^o_t$ the root velocity, and $\omega^o_t$ the root angular velocity.

\textbf{Network Initialization.} At the start of training, the humanoid already closely mimics the reference human motion, so the residual policy should ideally output values near zero. To enforce this, we initialize the final layer of the PPO actor using Xavier uniform initialization with a small gain factor (i.e., a scalar that scales the variance of the initialized weights) so that the initial outputs are close to zero~\cite{ankile2024imitationrefinementresidual}.

\begin{figure*}[htbp]
\centering
\includegraphics[width=0.99\linewidth]{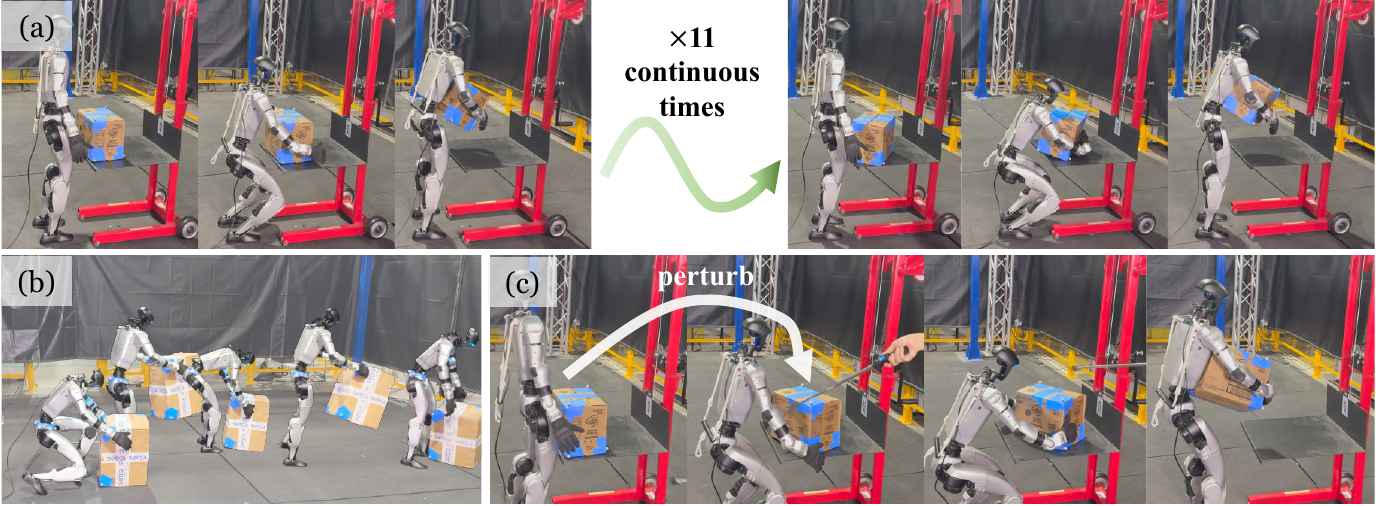}
\vspace{-0.05in}
\caption{We deploy \ourmethod on Unitree G1 with MoCap-based object states. (\textbf{a}) Lifting a box from random object initial poses across 11 trials; (\textbf{b}) Autonomous consecutive kneeling and box lifting; (\textbf{c}) Reactive behavior to external perturbations.
}
\vspace{-0.2in}
\label{fig/mocap_exp}
\end{figure*}

\textbf{Virtual Object Force Curriculum.} While the two-stage residual framework performs well on simple tasks with reliable reference trajectories and light objects, it often fails early when reference motions are noisy or objects are heavy. Failures arise mainly from (i) penetration introduced by kinematic retargeting, which causes the humanoid to push or knock over the object when naively imitating the reference--leading the initial policy to retreat rather than engage--and (ii) instability when handling large object masses. 

To address this, inspired by~\cite{mandi2025dexmachinafunctionalretargetingbimanual}, we introduce a virtual object controller curriculum that stabilizes training by driving the object toward its reference trajectory. At each timestep, PD controllers apply virtual forces and torques: 
$$\mathcal{F}_t = k_p (\hat{p}^o_t - p^o_t) - k_d v^o_t, \quad 
\mathcal{T}_t = k_p (\hat{\theta}^o_t \ominus \theta^o_t) - k_d \omega^o_t,$$
where $\mathcal{F}_t$ and $\mathcal{T}_t$ are the control force and torque, and $\ominus$ denotes the rotation difference. The controller gains $(k_p, k_d)$ are gradually decayed, so that early training is stabilized by strong virtual assistance, while later training forces the policy to take over and complete the task autonomously.

\subsubsection{Reward and Early Termination}  
Decoupling motion tracking from object interaction brings an additional advantage: we avoid carefully tuning the relative weights between motion and object rewards. Instead, we directly reuse the motion reward $r^m_t$ and domain randomization from GMT training, and introduce two additional terms: the object tracking reward $r^o_t$, which encourages task completion, and the contact tracking reward $r^c_t$, which provides explicit guidance on body–object contact, improving real-world deployability.

\textbf{Object Tracking Reward.} Prior work~\cite{liu2024opt2skill,xu2025intermimicuniversalwholebodycontrol} typically measures object tracking using pose differences between the simulated and reference objects, e.g.,  
$r^o_t = \exp(-\lambda_p \|p^o_t-\hat{p}^o_t\|_2) + w \cdot \exp(-\lambda_\theta \|\theta^o_t \ominus \hat{\theta}^o_t\|_2).$  
Instead, we propose an alternative with smoother reward landscape: sample $N$ points from the object mesh surface and compute the point-cloud difference between the current and reference states,
$$r^o_t = \exp(-\lambda_o \sum^N_{i=1}\|\mathbf{P}[i]_t - \hat{\mathbf{P}}[i]_t\|_2),$$  
where $\mathbf{P}_t \in \mathbb{R}^{N \times 3}$ denotes the sampled 3D points. This approach naturally accounts for both translation and rotation, eliminating the need for task-specific weight tuning. 

\textbf{Contact Reward.} To encourage correct physical interactions during whole-body manipulation while remaining efficient, we discretize contact locations into meaningful links, e.g., torso, hip, and arms, excluding feet since they mainly contact the ground. Oracle contact information is obtained from the reference human–object interaction trajectory:  
$$\hat{c}_t[i] = \mathbf{1}(\|\hat{d}_t[i]\| < \sigma_c),$$  
where $i$ indexes the links, $\mathbf{1}(\cdot)$ is the indicator function, and $\|\hat{d}_t[i]\|$ is the distance between link $i$ and the object surface. The contact tracking reward is then defined as  
$$r^c_t = \sum_i \hat{c}_t[i] \cdot \exp\Big(-\frac{\lambda}{f_t[i]}\Big),$$  
where $f_t[i]$ is the contact force at link $i$.

\textbf{Early Termination.} Commonly used in motion tracking, early termination~\cite{2018-TOG-deepMimic} ends an episode if a body part makes unintended ground contact or deviates substantially from the reference, preventing the policy from overvaluing invalid states. For humanoid whole-body loco-manipulation, we introduce additional conditions:  
(i) the object mesh deviates from its reference beyond a threshold, $\|\mathbf{P}_t-\hat{\mathbf{P}}_t\|_2 > \sigma_o$, or  
(ii) any required body–object contact is lost for more than $10$ consecutive frames.

%% file: sections/exp.tex
\input{tables/main_sim_exp}
\section{EXPERIMENTS}

\begin{figure*}[htbp]
\centering
\includegraphics[width=0.99\linewidth]{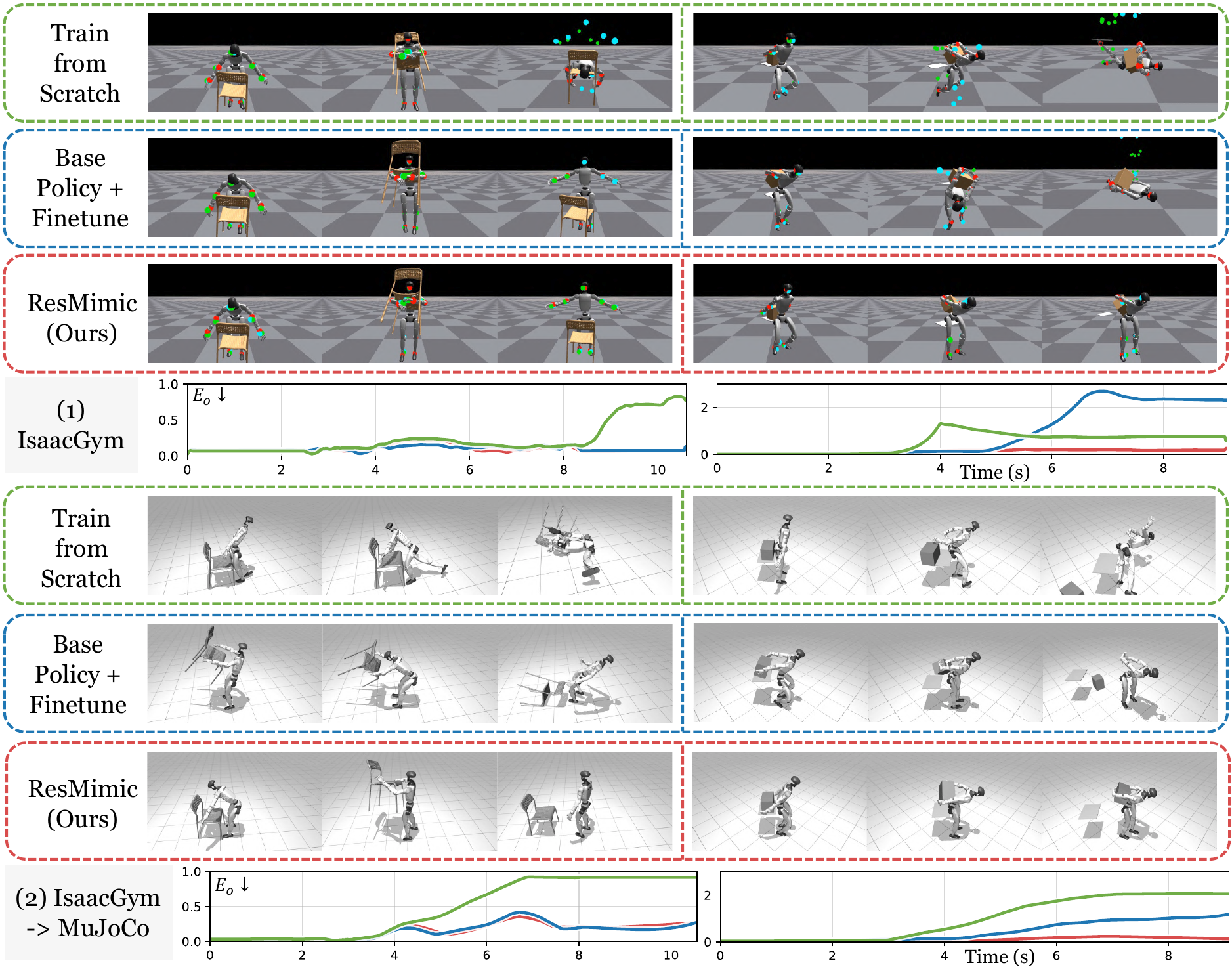}
\vspace{-0.05in}
\caption{
Comparison between IsaacGym and MuJoCo results for task \textit{Chair} (left) and \textit{Carry} (right). 
Corresponding curves quantify object tracking error for \textcolor[HTML]{70AD47}{Train from Scratch}, \textcolor[HTML]{1F77B4}{Finetune}, and
\textcolor[HTML]{D94D4D}{ResMimic}.
}
\vspace{-0.2in}
\label{fig/main_sim_exp}
\end{figure*}

We assess the effectiveness of \ourmethod through a combination of large-scale simulation studies and real-world deployment on a Unitree G1 humanoid robot (29-DoF, 1.3\,m tall). 

The evaluation is designed to investigate both algorithmic efficiency and deployment robustness. 
In particular, we focus on the following research questions:

\textbf{Q1:} Can a general motion tracking (GMT) policy, without task-specific retraining, accomplish diverse loco-manipulation tasks?  

\textbf{Q2:} Does initializing from a pre-trained GMT policy improve training efficiency and final performance compared to training from scratch?  

\textbf{Q3:} When adapting GMT policies to loco-manipulation tasks, is residual learning more effective than fine-tuning?  

\textbf{Q4:} Beyond simulation, can \ourmethod achieve precise, expressive, and robust control in the real-world?

\subsection{Experiment Setup}
\subsubsection{Tasks}
We design four challenging whole-body loco-manipulation tasks that stress different aspects of humanoid control and generalization. Task (i) \textit{Kneel on one knee and lift a box} — requires expressive, large-amplitude motion and precise lower-body coordination.  (ii) \textit{Carry a box onto the back} — demands whole-body expressiveness while maintaining balance under shifting load distribution.  
(iii) \textit{Squat and lift a box with arms and torso} — highlights the challenge of whole-body contact-rich manipulation.
(iv) \textit{Lift up a chair} — involves manipulating a heavy, irregularly shaped object. For clarity, we will refer to these tasks as \textit{Kneel}, \textit{Carry}, \textit{Squat}, and \textit{Chair}, respectively.
The human–object interaction reference motions for these tasks are collected using OptiTrack motion capture system.

\subsubsection{Evaluation Metrics}
We evaluate \ourmethod in terms of training efficiency, motion fidelity, manipulation accuracy, and overall task completion. 
The following metrics are used throughout our experiments:
(i) \textbf{Training Iterations (\textit{Iter.})}: To eliminate discrepancies due to hardware differences, we report convergence speed in terms of training iterations rather than wall-clock time. Here we determine convergence when reward stops increasing approximately.  
(ii) \textbf{Object Tracking Error}: 
        $E_o = \frac{1}{T}\sum_{t=1}^T \sum^N_{i=1}|| \mathbf{P}[i]_t - \hat{\mathbf{P}}[i]_t ||_2,$
    where $\mathbf{P}_t$ and $\hat{\mathbf{P}}_t$ denote sampled point clouds on the manipulated object mesh and its reference at time $t$. 
(iii) \textbf{Motion Tracking Error}:  
    $E_m = \frac{1}{T}\sum_{t=1}^T \sum_i || p_t[i] - \hat{p}_t[i] ||_2,$
    where $p_t[i]$ and $\hat{p}_t[i]$ are the global positions of link $i$ at time $t$.   
(iv) \textbf{Joint Tracking Error}:  
$
        E_j = \frac{1}{T}\sum_{t=1}^T || q_t - \hat{q}_t ||_2,
$
    where $q_t, \hat{q}_t \in \mathbb{R}^{29}$ denote the robot’s joint angles and their reference values. This evaluates joint-level precision.  
(v) \textbf{Task Success Rate (\textit{SR})}: A rollout is considered successful if $E_o$ is below a predefined threshold and the robot remains balanced.

\subsubsection{Baselines}
To validate the effectiveness and efficiency of our pipeline, we compare \ourmethod against three representative and strong baselines:

\begin{enumerate}[label=(\roman*)]
    \item \textbf{Base Policy}: The pre-trained GMT policy is directly deployed to follow human reference motion, without access to object information.  
    \item \textbf{Train from Scratch}: A single-stage RL policy is trained from scratch to track both human motion and object trajectories, without leveraging the GMT policy. For fairness, we use the same reward terms as \ourmethod across all tasks, without task-specific tuning.  
    \item \textbf{Base Policy + Fine-tune}: The base GMT policy is fine-tuned to track both human motion and object trajectories. The reward terms are identical to those used in \ourmethod. However, due to the limitations of direct fine-tuning, the policy cannot incorporate explicit object information as input.  
\end{enumerate}

\subsection{Sim-to-Sim Evaluation}

All policies are trained in IsaacGym to leverage its massive parallelism for accelerated data collection. 
To evaluate generalization, however, we perform sim-to-sim transfer into MuJoCo, which is widely regarded as a better proxy for real-world physics and thus a closer benchmark for sim-to-real performance. 
We report quantitative results for all baselines across the four loco-manipulation tasks in Table~\ref{tab/main_exp}, and highlight key takeaways below.  

\textbf{(Q1) GMT alone cannot complete loco-manipulation tasks, but provides a strong initialization.}  
We first compare \ourmethod against directly deploying the pre-trained GMT policy. 
As shown in Table~\ref{tab/main_exp}, the base GMT policy achieves only a $10\%$ success rate, compared to $92.5\%$ for \texttt{ResMimic}. 
Although GMT alone yields slightly lower joint tracking error (due to its training objective), it performs poorly on object tracking and overall task completion, as it lacks access to object information. 
This indicates that while GMT captures joint-level precision, it is insufficient for manipulation without adaptation.  

\begin{figure}[!t]
\centering
\includegraphics[width=0.99\linewidth]{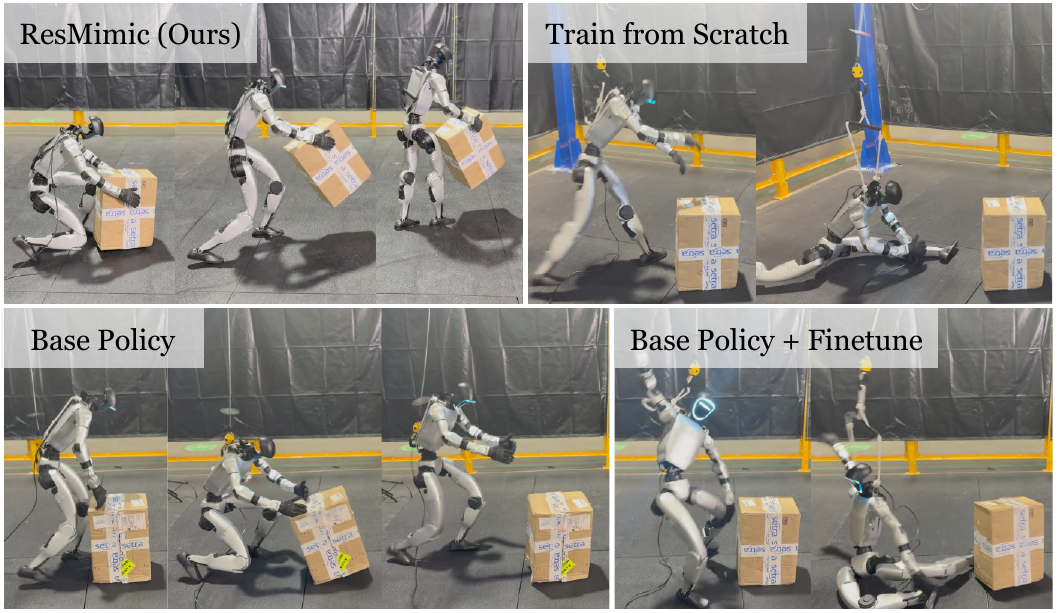}
\caption{Real-world qualitative results comparing \ourmethod against all other baselines.
}
\vspace{-0.25in}
\label{fig/real_comparison}
\end{figure}

\textbf{(Q2) Using GMT as base policy significantly improves training efficiency and effectiveness.}  
Next, we compare \ourmethod with policies trained entirely from scratch under identical settings (reward function, domain randomization, etc.). 
As shown in Table~\ref{tab/main_exp}, training from scratch fails to solve the tasks in MuJoCo and converges much more slowly. 
A side-by-side comparison between IsaacGym and MuJoCo as shown in Figure~\ref{fig/main_sim_exp} reveals that trained-from-scratch policies sometimes show partial success in IsaacGym but collapse entirely under sim-to-sim transfer. 
In contrast, \ourmethod maintains strong performance with minimal degradation. 
This demonstrates the necessity of using GMT as a foundation: its large-scale pretraining imbues generalization and robustness against sim-to-sim gaps.

\textbf{(Q3) Residual learning outperforms direct fine-tuning.}  
Finally, we evaluate residual learning against fine-tuning the GMT policy. 
While fine-tuning yields slight improvements over training from scratch, it neither outperforms the base GMT policy nor approaches the performance of \ourmethod as shown in Table~\ref{tab/main_exp}. 
A key limitation is that fine-tuning cannot incorporate additional object observations, since the GMT policy architecture is restricted to human motion inputs. 
Although object-tracking rewards provide some supervision, the lack of explicit object state prevents learning robust behaviors, particularly under randomized object poses. 
Moreover, fine-tuning tends to overwrite the generalization capability of the GMT policy, leading to instability across tasks. 
A comparison between IsaacGym and MuJoCo as shown in Figure~\ref{fig/main_sim_exp} reveals that fine-tuned policies succeed on task \textit{Chair-Lift} in IsaacGym but fail to transfer in MuJoCo, underscoring the superiority of residual learning as a more generalizable and extensible adaptation strategy.

\begin{figure}[!t]
\centering
\includegraphics[width=0.99\linewidth]{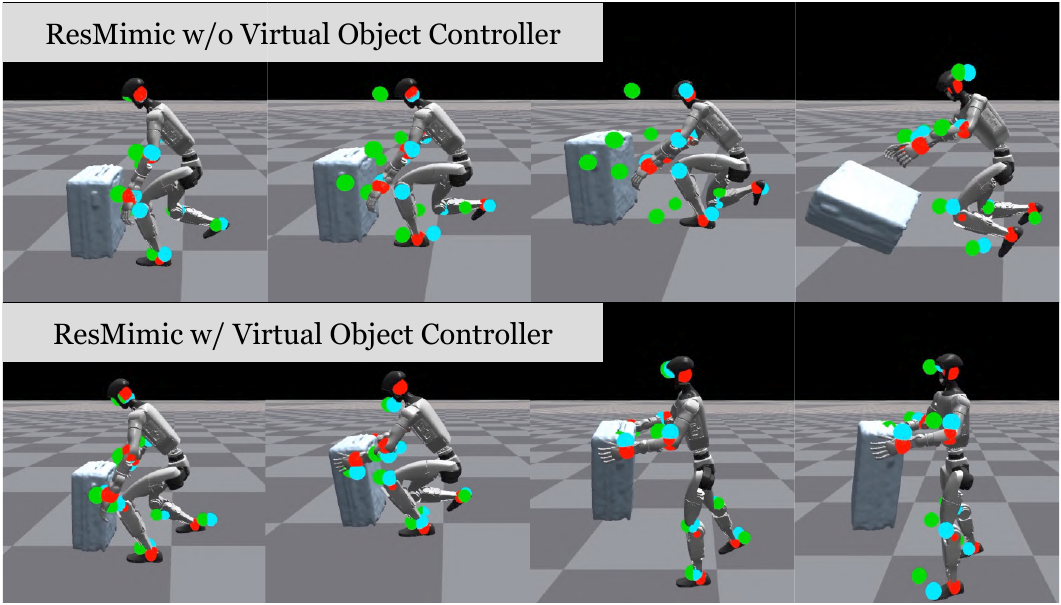}
\caption{{Ablation on virtual object controller.}
}
\label{fig/voc_ablation}
\vspace{-0.2in}
\end{figure}

\subsection{Real-world Evaluation}

As shown in Figure~\ref{fig/teaser}, we deploy \ourmethod on a Unitree G1 humanoid and demonstrate precise, expressive, robust whole-body loco-manipulation. \ourmethod supports both blind (without object state input) and non-blind deployment (with MoCap-based object state input). For simplicity, all real-world results in Figure~\ref{fig/teaser} are under blind deployment.

\begin{itemize}
    \item \textbf{Expressive carrying motions:} the robot kneels on one knee to pick up a box, or carry the box on its back, highlighting expressive whole-body movement.
    \item \textbf{Humanoid--object interaction beyond manipulation:} the robot sits down on a chair and then stands up while maintaining balance and contact with the environment.
    \item \textbf{Heavy payload carrying with whole-body contact:} the robot successfully carries a 4.5\,kg box, while the G1’s wrist payload limit is around 2.5\,kg, demonstrating the necessity of leveraging whole-body contact.
    \item \textbf{Generalization to irregular heavy objects:} the robot lifts and carries chairs weighing 4.5\,kg and 5.5\,kg, showing instance-level generalization to novel, non-box geometries.
\end{itemize}

We also conduct a qualitative comparison of \ourmethod against all baselines in the real world, as shown in Fig.~\ref{fig/real_comparison}. The results indicate that while the base policy can superficially mimic human motion, it lacks object awareness—an issue that becomes even more pronounced when the demonstration data is imperfect.
Training from scratch and finetuning, on the other hand, fail entirely due to sim-to-real gap.

Finally, we evaluate \ourmethod under non-blind deployment with MoCap-based object state input as shown in Figure~\ref{fig/mocap_exp}. In this setting, the robot demonstrates the ability to (i) manipulate objects from random initial poses, (ii) autonomously perform consecutive loco-manipulation tasks, and (iii) show reactive behavior to external perturbations.

\subsection{Ablation Studies}
\begin{figure}[!t]
\centering
\includegraphics[width=0.99\linewidth]{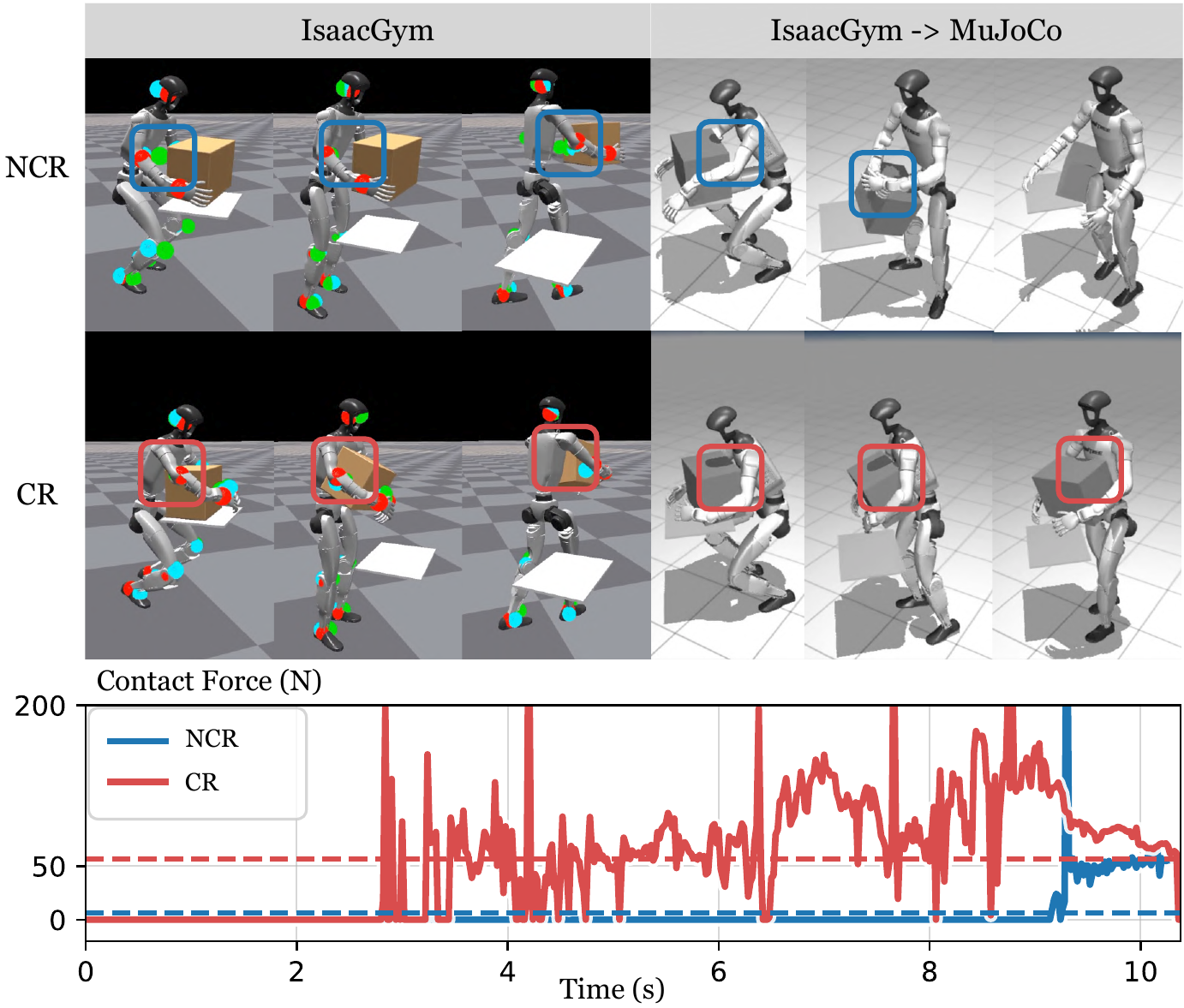}
\caption{\textbf{Ablation on contact reward.} Here NCR denotes ``No Contact Reward'', and CR denotes ``with Contact Reward''. Corresponding curves (bottom) quantify torso contact force.}
\label{fig/contact_ablation}
\vspace{-0.2in}
\end{figure}
\subsubsection{Effect of the Virtual Object Controller}
The virtual object controller stabilizes early-stage training by applying curriculum-based virtual forces that guide the object toward its reference trajectory. A qualitative example is shown in Figure~\ref{fig/voc_ablation}. In this task, the reference motions contain imperfections, including penetrations between the humanoid’s hand and the object. The policy initially focuses only on motion tracking to reach the object, causing the object to be knocked over, which yields low object rewards and frequent early terminations. This quickly drives the policy into a local minimum where the robot retreats rather than engaging with the object. In contrast, with the virtual force curriculum, the object remains stabilized during early learning, enabling the policy to overcome motion-data imperfections and converge to precise manipulation strategies.

\subsubsection{Effect of the Contact Reward}
The contact reward provides explicit guidance on leveraging whole-body strategies. As illustrated in Figure~\ref{fig/contact_ablation}, there are two possible ways to lift the box: (1) relying only on wrists and hands, or (2) engaging both torso and arm contact as demonstrated by humans. Without the contact reward, the policy converges to (1), which may succeed in IsaacGym but fails to transfer to MuJoCo and the real world. With the contact reward, the humanoid instead adopts strategy (2), using coordinated torso and arm contact. This alignment with human demonstrations results in improved sim-to-sim and sim-to-real transfer, validating the importance of the contact reward.

%% file: tables/main_sim_exp.tex
\begin{table}[htbp]
\caption{Sim-to-Sim evaluation of \ourmethod against baseline approaches in MuJoCo across all four tasks.}
\scriptsize
\label{tab/main_exp}
\renewcommand{\arraystretch}{1.2}
\setlength{\tabcolsep}{4pt}
\begin{center}
\begin{tabular}{c|c|c|c|c|c|c}
\hline
\rowcolor{gray!10}
Method & Task & \textit{SR} $\uparrow$ & \textit{Iter.} $\downarrow$ & $E_{o}$ $\downarrow$ & $E_{m}$ $\downarrow$ & $E_{j}$ $\downarrow$ \\
\hline
\multirow{5}{*}{\shortstack{Base \\ Policy}} 
    & Kneel &  0\% & $-$ & 0.76 $\pm$ 0.01  & 3.30 $\pm$ 0.53& 0.28 $\pm$ 0.01\\
   & Carry & 0\%  & $-$ & 0.29 $\pm$ 0.02 & 2.47 $\pm$ 0.26 & 1.19 $\pm$ 0.30 \\
   & Squat & 40\% & $-$ & 0.19 $\pm$ 0.01 & 0.93 $\pm$ 0.07 & 0.90 $\pm$ 0.08 \\ 
   & Chair & 0\% & $-$ & 1.19 $\pm$ 0.48 & 30.18 $\pm$ 33.45 & 1.20 $\pm$ 0.23  \\ \cline{2-7}
   &  \cellcolor{gray!40} Mean & \cellcolor{gray!40}{10\%} & \cellcolor{gray!40}$-$ &\cellcolor{gray!40} 0.61 &\cellcolor{gray!40} 9.22 & \cellcolor{gray!40}{\textbf{0.89 }} \\
\hline
\multirow{5}{*}{\shortstack{Train\\ from \\ Scratch}} 
    & Kneel & 0\% & $\times$ & 0.69 $\pm$ 0.00 & 5.20 $\pm$ 0.62 & 3.41 $\pm$ 0.07 \\
   & Carry & 0\% & 6500 & 0.70 $\pm$ 0.03  & 5.39 $\pm$ 0.38 &  2.33 $\pm$ 0.06  \\
   & Squat & 0\% & 5000 & 0.68 $\pm$ 0.05& 7.56 $\pm$ 2.31 & 4.28 $\pm$ 0.70 \\
   & Chair & 0\% & 2000 & 0.97 $\pm$ 0.08 & 10.01 $\pm$ 1.28 & 13.36 $\pm$ 0.92 \\ \cline{2-7} 
   & \cellcolor{gray!40} Mean & \cellcolor{gray!40}0\% & \cellcolor{gray!40}4500 & \cellcolor{gray!40}0.76 & \cellcolor{gray!40}7.04 & \cellcolor{gray!40}5.84 \\
\hline
\multirow{5}{*}{\shortstack{Finetune}} 
    & Kneel & 0\% & $\times$ & 0.87 $\pm$ 0.01 & 5.92 $\pm$ 0.81  & 3.02 $\pm$ 0.18 \\
   & Carry & 30\% & 4500 & 0.33 $\pm$ 0.01 & 2.49 $\pm$ 0.18 & 2.39 $\pm$ 0.06 \\
   & Squat & 0\% & 2000 & 0.47 $\pm$ 0.05 & 5.07 $\pm$ 1.06 & 2.53 $\pm$ 0.13 \\
   & Chair & 0\% & 700 &  0.15 $\pm$ 0.01 & 0.28 $\pm$ 0.05  & 1.26 $\pm$ 0.09 \\ 
   \cline{2-7}
   &  \cellcolor{gray!40}Mean & \cellcolor{gray!40}7.5\% & \cellcolor{gray!40}{2400} & \cellcolor{gray!40}{0.46} & \cellcolor{gray!40}{3.44} &  \cellcolor{gray!40}2.30\\
\hline
\multirow{5}{*}{\shortstack{ResMimic\\ (Ours)}} 
    & Kneel & 90\% & 2000 & 0.14 $\pm$ 0.00 & 0.23 $\pm$ 0.06 & 2.17 $\pm$ 0.06\\
   & Carry & 100\% & 1000 & 0.11 $\pm$ 0.00 & 0.08 $\pm$ 0.00 & 1.24 $\pm$ 0.03 \\
   & Squat & 80\% & 1500 & 0.07 $\pm$ 0.01 & 0.07 $\pm$ 0.03 & 1.18 $\pm$ 0.03  \\ 
   & Chair & 100\% & 700 & 0.16 $\pm$ 0.01 & 0.13 $\pm$ 0.02 & 0.55 $\pm$ 0.01 \\ \cline{2-7}
   &  \cellcolor{gray!40}Mean & \cellcolor{gray!40}\textbf{\textbf{92.5\%}} & \cellcolor{gray!40}\textbf{\textbf{1300}} & \cellcolor{gray!40}\textbf{\textbf{0.12}} & \cellcolor{gray!40}\textbf{\textbf{0.13}} & \cellcolor{gray!40}{1.29} \\
\hline
\end{tabular}
\end{center}
\vspace{-0.2in}
\end{table}

%% file: sections/conclusion.tex
\section{CONCLUSION}
This work introduces \ourmethod, a two-stage residual learning framework that enables precise, expressive, and robust humanoid loco-manipulation. We first pre-train a general motion-tracking policy on large-scale human motion data, and then refine it with a task-specific residual policy. Across extensive experiments, \ourmethod delivers substantial gains in task success, motion fidelity, and training efficiency, and further demonstrates seamless deployment on a real Unitree G1 humanoid. These results highlight the transformative potential of pre-trained policies for humanoid control.